\documentclass[10pt,english,twocolumn]{article}
\usepackage[latin9]{inputenc}
\usepackage{array}
\usepackage{booktabs}
\usepackage{amsmath}
\usepackage{amssymb}
\usepackage{graphicx}

\makeatletter

\providecommand{\tabularnewline}{\\}

\usepackage{latex12}
\usepackage{times}
\usepackage{balance}
\usepackage{epstopdf}

\makeatother

\usepackage{babel}
\begin{document}

\title{Stabilizing Sparse Cox Model using Clinical Structures in Electronic
Medical Records}

\author{Shivapratap Gopakumar, Truyen Tran, Dinh Phung, Svetha Venkatesh\\
 \emph{Center for Pattern Recognition and Data Analytics}, Deakin
University, Australia\\
 \emph{\{sgopakum, truyen.tran, dinh.phung, svetha.venkatesh\}@deakin.edu.au}}
\maketitle
\begin{abstract}
Stability in clinical prediction models is crucial for transferability
between studies, yet has received little attention. The problem is
paramount in high-dimensional data which invites sparse models with
feature selection capability. We introduce an effective method to
stabilize sparse Cox model of time-to-events using clinical structures
inherent in Electronic Medical Records (EMR). Model estimation is
stabilized using a feature graph derived from two types of EMR structures:
temporal structure of disease and intervention recurrences, and hierarchical
structure of medical knowledge and practices. We demonstrate the efficacy
of the method in predicting time-to-readmission of heart failure patients.
On two stability measures -- the Jaccard index and the Consistency
index -- the use of clinical structures significantly increased feature
stability without hurting discriminative power. Our model reported
a competitive AUC of 0.64 (95\% CIs: {[}0.58,0.69{]}) for 6 months
prediction. 

\global\long\def\bx{\boldsymbol{x}}

\global\long\def\bw{\boldsymbol{w}}

\end{abstract}

\section{Introduction}

Heart failure is a serious illness which demands frequent hospitalization.
It is estimated that 50\% of heart failure patients are readmitted
within 6 months after their discharge \cite{desai2012rehospitalization}.
A significant amount of these readmissions can be predicted and prevented.
Existing readmission prediction models use diverse subsets of clinical
variables from prior hypotheses or medical literature \cite{RossJSMulveyGK2008},
making it hard to reach a consensus of what are predictive and what
are not. With the emergence of electronic medical records (EMR), it
is now possible to obtain data on all aspects of patient care over
time. A typical EMR database contains full details about demographics,
history of hospital visits, diagnosis, procedures, physiological measurements,
bio-markers and interventions that are recorded over time \cite{jensen2012mining}.
Such high-dimensional data is comprehensive but calls for robust feature
selection when deriving prediction models. Unfortunately, automatic
feature selection, particularly in clinical data, has been known to
cause instability in features against data sampling, and thus limiting
the reproducibility of the model \cite{austin2004automated}. Improving
stability of model estimation is crucial, but it has received little
attention.

We investigate time-to-readmission due to heart failure using a sparsity-inducing
Cox model \cite{tibshirani1997lasso} on high-dimensional EMR data.
Our objective is to improve the stability of this sparse model by
exploiting clinical structures inherent in the EMR data. Specifically,
we make use of (i) temporal relations in diagnosis, prognosis and
intervening events, and (ii) hierarchical structures of disease family
through semantics in ICD-10 tree%
\footnote{http://apps.who.int/classifications/icd10%
} and procedure pool (ACHI)%
\footnote{https://www.aihw.gov.au/procedures-data-cubes/%
} . These structures are encoded into a feature graph with its edges
representing the relation between features. The application of feature
graph ensures sharing of statistical strength among correlated features,
thereby promoting stability. The proposed stabilized sparse Cox model
is trained and validated using retrospective data from a local hospital
in Australia. Model stability is validated using measures of Jaccard
index \cite{real1996probabilistic} and Consistency index \cite{kuncheva2007stability}.
We demonstrate that by exploiting inherent clinical structures in
EMR, the stability of our regularized Cox model is improved without
loss in performance.

\section{Method}

We describe a method to utilize inherent structures in data to stabilize
sparse Cox model derived from EMR databases. To start with, we employed
the one-sided convolutional filter bank introduced in \cite{TranPLHBV13}
to extract a large pool of features from EMR databases. The filter
bank summarizes event statistics over multiple time periods and granularities.

\subsection{Sparse Cox Model}

We use Cox regression to model risk of readmission due to heart failure
(hazard function) at a future instant, based on data from EMR. Unlike
logistic regression where each patient is assigned a nominal label,
Cox regression models the readmission time directly \cite{Vinzamuri2013cox}.
The proportional hazards assumption in Cox regression assumes a constant
relationship between readmission time and EMR-derived explanatory
variables. Let $\mathcal{D}=\left\{ \bx_{\ell},y_{\ell}\right\} _{\ell=1}^{n}$
be the training dataset, ordered on increasing $y_{\ell}$, where
$\bx_{\ell}\in\mathbb{R}^{p}$ denotes the feature vector for $\ell^{th}$
index admission and $y_{\ell}$ is the time to next unplanned readmission.
When a patient withdraws from the hospital or does not encounter readmission
in our data during the follow-up period, the observation is treated
as right censored. Let $q$ observations be uncensored and $R(t_{\ell})$
be the remaining events at readmission time $t_{\ell}$.

Since the data $\mathcal{D}$ is high dimensional (possibly $p\gg n$),
we apply lasso regularization for sparsity induction \cite{tibshirani1997lasso}.
The feature weights $\boldsymbol{w}\in\mathbb{R}^{p}$ are estimated
by maximizing the $\ell_{1}$-penalized partial likelihood: 

\begin{equation}
\mathcal{L}_{1}^{reg}=\frac{1}{n}\mathcal{L}\left(\boldsymbol{w};\mathcal{D}\right)-\alpha\Vert\boldsymbol{w}\Vert_{1}\label{eq:lasso_cox}
\end{equation}
where $\Vert\boldsymbol{w}\Vert_{1}=\sum_{i}\left|w_{i}\right|$,
$\alpha>0$ is the regularizing constant, and $\mathcal{L}\left(\mathbf{w};\mathcal{D}\right)$
is the log partial likelihood \cite{cox1975partial} computed as:
\[
\overset{_{q}}{\underset{_{\ell=1}}{\sum}}\left\{ \boldsymbol{w}^{\top}\bx_{\ell}-\log\left[\underset{\jmath\in R(t_{\ell})}{\sum}\exp\left(\boldsymbol{w}^{\top}\bx_{\jmath}\right)\right]\right\} 
\]

The lasso induces sparsity by driving the weights of weak features
towards zero. However, sparsity induction is known to cause instability
in feature selection \cite{xu2012sparse}. Instability occurs because
lasso randomly chooses one in two highly correlated features. Each
training run with slightly different data could result in a different
feature from the correlated pair. The nature of EMR data further aggravates
this problem. The EMR data is, by design, highly correlated and redundant.
Also, features in the EMR data maybe weakly predictive for some task,
thereby limiting the probability that they are selected.  These sum
up to lack of reproducibility between model updates or external validations,
hindering the method credibility and adoption by clinicians.

\subsection{Stabilizing with Clinical Structures}

A natural solution to the instability problem is to ensure that correlated
features are assigned with similar weights. We exploited two clinical
structures inherent in the EMR data for this purpose. The first is
\emph{temporal structure }of diagnosis, hospital interaction and intervening
events recorded over time. The second is \emph{code structure} based
on hierarchies in medical knowledge and practices such as International
Classification of Disease, $10^{th}$ revision (ICD-10) and procedure
codes. Two codes are considered to be correlated if they share the
same prefix. Using the temporal structure of events, and the hierarchical
structure of code trees, we built an undirected graph with features
as nodes and edges representing the relation between features. Let
$\mathbf{A}\in\mathbb{R}^{p\times p}$ be the incidence matrix of
the feature graph with $A_{ij}=1$ if features $i$ and $j$ share
a temporal or code relation, and $A_{ij}=0$ otherwise. We introduced
a graph regularizing term to Eq. (\ref{eq:lasso_cox}):
\begin{equation}
\mathcal{L}_{2}^{reg}=\mathcal{L}_{1}^{reg}-\frac{1}{2}\beta\underset{_{ij}}{\sum}A_{ij}(w_{i}-w_{j})^{2}\label{eq:graph_lasso_cox_fn}
\end{equation}
where the term $\underset{_{ij}}{\sum}A_{ij}(w_{i}-w_{j})^{2}$ ensures
similar weights for correlated features, and $\beta>0$ is the correlation
coefficient.

The graph regularization term $\sum_{ij}A_{ij}(w_{i}-w_{j})^{2}$
can be expressed as:$\underset{_{i}}{\sum}\left(\underset{_{j}}{\sum}A_{ij}\right)w_{i}^{2}-\underset{_{ij}}{\sum}A_{ij}w_{i}w_{j}$.
Let $\mathbf{L}$ denote the Laplacian of feature graph $\mathbf{A}$,
where $L_{ii}=\Sigma_{j}A_{ij}$ and $L_{ij}=-A_{ij}$ \cite{chung1997spectral},
the graph regularization term now becomes $\boldsymbol{w}^{\top}\mathbf{L}\boldsymbol{w}$.
Eq. (\ref{eq:graph_lasso_cox_fn}) can be rewritten as:

\begin{equation}
\mathcal{L}_{2}^{reg}=\frac{1}{n}\mathcal{L}\left(\boldsymbol{w};\mathcal{D}\right)-\alpha||\boldsymbol{w}||_{1}-\frac{1}{2}\beta\,\boldsymbol{w}^{\top}\mathbf{L}\boldsymbol{w}\label{eq:final_graph_regularized_cox}
\end{equation}

The gradient of Eq.(\ref{eq:final_graph_regularized_cox}) becomes:

\begin{eqnarray}
\frac{\partial\mathcal{L}_{2}^{reg}}{\partial\bw} & = & \overset{_{q}}{\underset{_{\ell=1}}{\sum}}\left\{ \bx_{(\ell)}-\frac{\underset{\jmath\in R(t_{\ell})}{\sum}\bx_{\jmath}\exp(\boldsymbol{w}^{\top}\bx_{\jmath})}{\underset{\jmath\in R(t_{\ell})}{\sum}\exp(\boldsymbol{w}^{\top}\bx_{\jmath})}\right\} \nonumber \\
 &  & \quad\quad-\alpha\,\mbox{sign}(\bw)-\beta\bw^{\top}\mathbf{L}\label{eq:reg_cox_gradient}
\end{eqnarray}

\subsection{Measuring Model Stability}

We used the Jaccard index \cite{real1996probabilistic} and the Consistency
index \cite{kuncheva2007stability} to measure stability of feature
selection process. To simulate data variations due to sampling, we
created $B$ data bootstraps of original size $n$. For each bootstrap,
a model was trained and a subset of top $k$ features was selected.
Features were ranked according to their importance, which is product
of feature weight and standard deviation. Finally, we obtained a list
of feature subsets $S=\left\{ S_{1},S_{2},...,S_{B}\right\} $ where
$\left|S_{b}\right|=k$.

The \emph{Jaccard index} measures similarity as a fraction between
cardinalities of intersection and union of feature subsets. Given
two feature sets $S_{a}$ and $S_{b}$, the pairwise Jaccard index
reads:

\begin{eqnarray}
J_{C}(S_{a},S_{b}) & = & \frac{\left|S_{a}\cap S_{b}\right|}{\left|S_{a}\cup S_{b}\right|}\label{eq:pairwise_Jaccard_index}
\end{eqnarray}
The \emph{Consistency index} corrects the overlapping due to chance.
Considering a pair of subsets $S_{i}$ and $S_{j}$, the pairwise
Consistency index $I_{C}$ is defined as:

\begin{equation}
I_{C}(S_{a},S_{b})=\frac{rd-k^{2}}{k(d-k)}\label{eq:pairwise_consistency_index}
\end{equation}
in which $\left|S_{a}\cap S_{b}\right|=r$ and $d$ is the number
of features. The stability for the set $S=\left\{ S_{1},S_{2},...,S_{B}\right\} $
is calculated as average across all pairwise $J_{C}(S_{a},S_{b})$
and $I_{C}(S_{a},S_{b})$. Jaccard index is bounded in $\left[0,1\right]$
while Consistency index is bounded in $\left[-1,+1\right]$.

\section{Results}

We trained our model on 1,088 patients (1,405 index admissions) discharged
from Barwon Health (a regional hospital in Australia) from Jan 2007
to Sept 2010. The model was validated on another cohort of 317 patients
(369 index admissions) discharged from Oct 2010 to Dec 2011. From
a total of 3,338 features, our model selected 94 features that are
highly predictive of heart failure readmissions. The top predictors
are listed in Table \ref{tab:The-top-predictors}.

\begin{figure*}[tp]
\begin{centering}
\begin{tabular}{ccc}
\includegraphics[width=0.3\textwidth,height=0.25\textwidth]{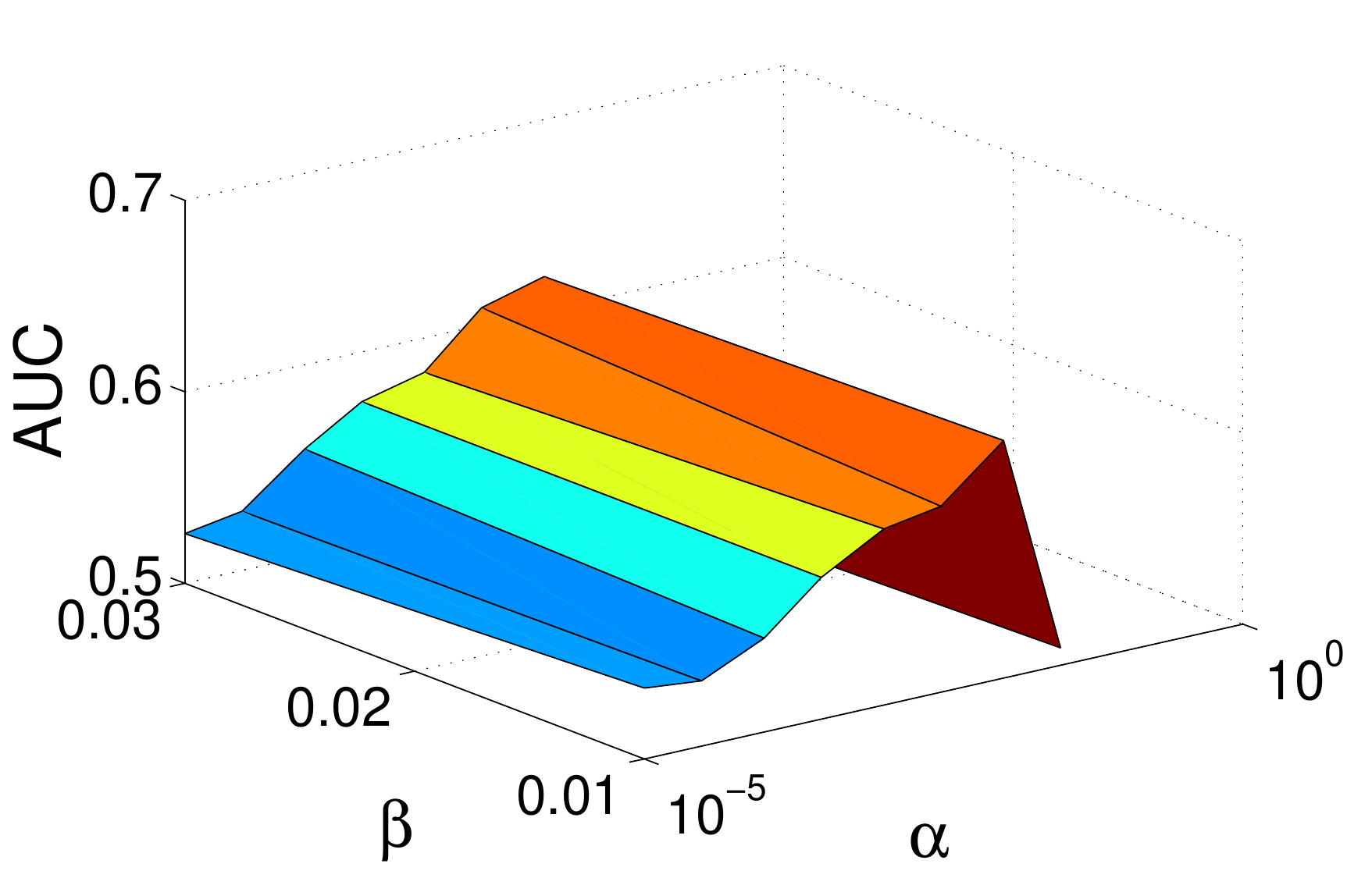} & \includegraphics[width=0.3\textwidth,height=0.25\textwidth]{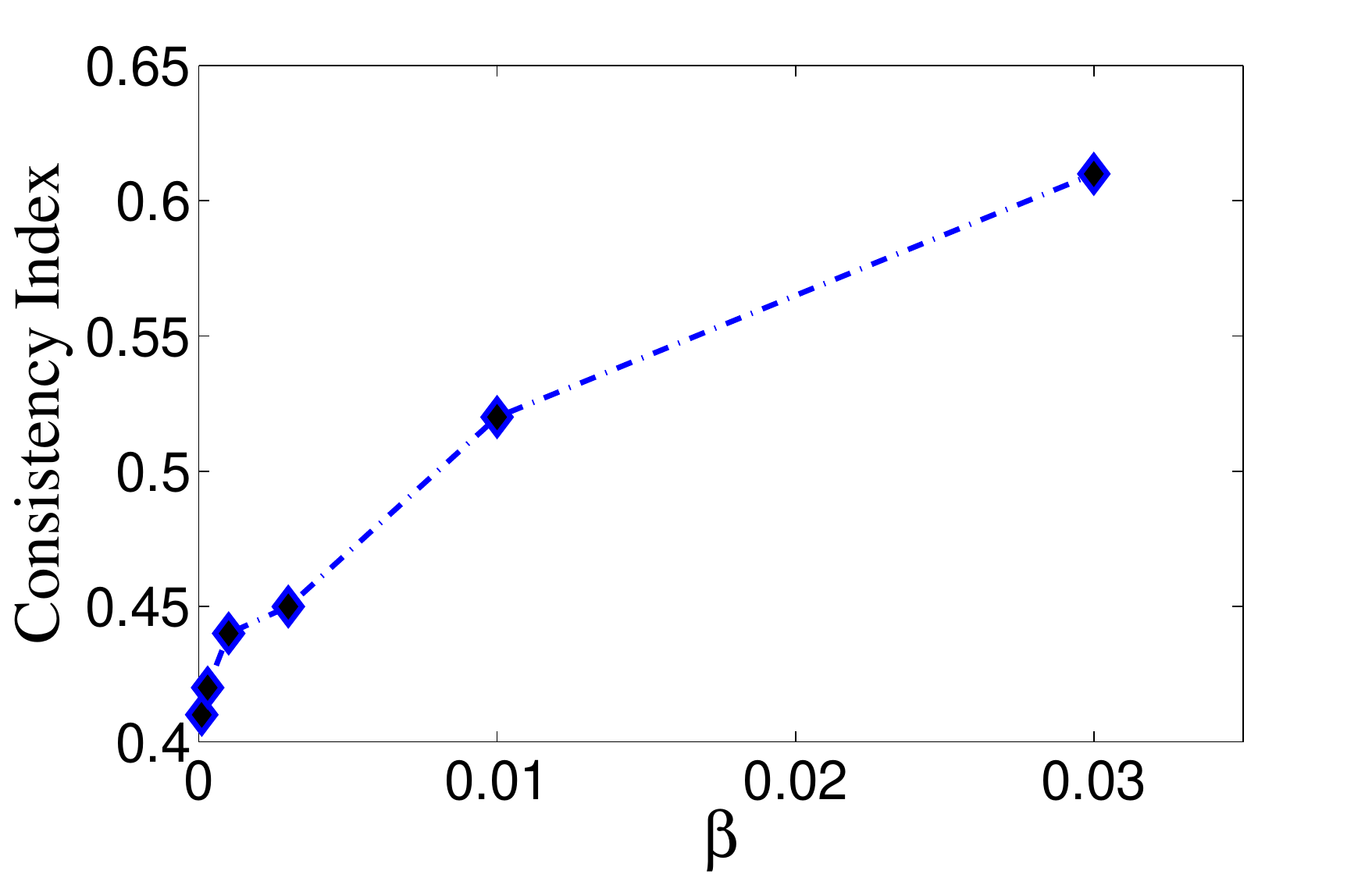} & \includegraphics[width=0.3\textwidth,height=0.25\textwidth]{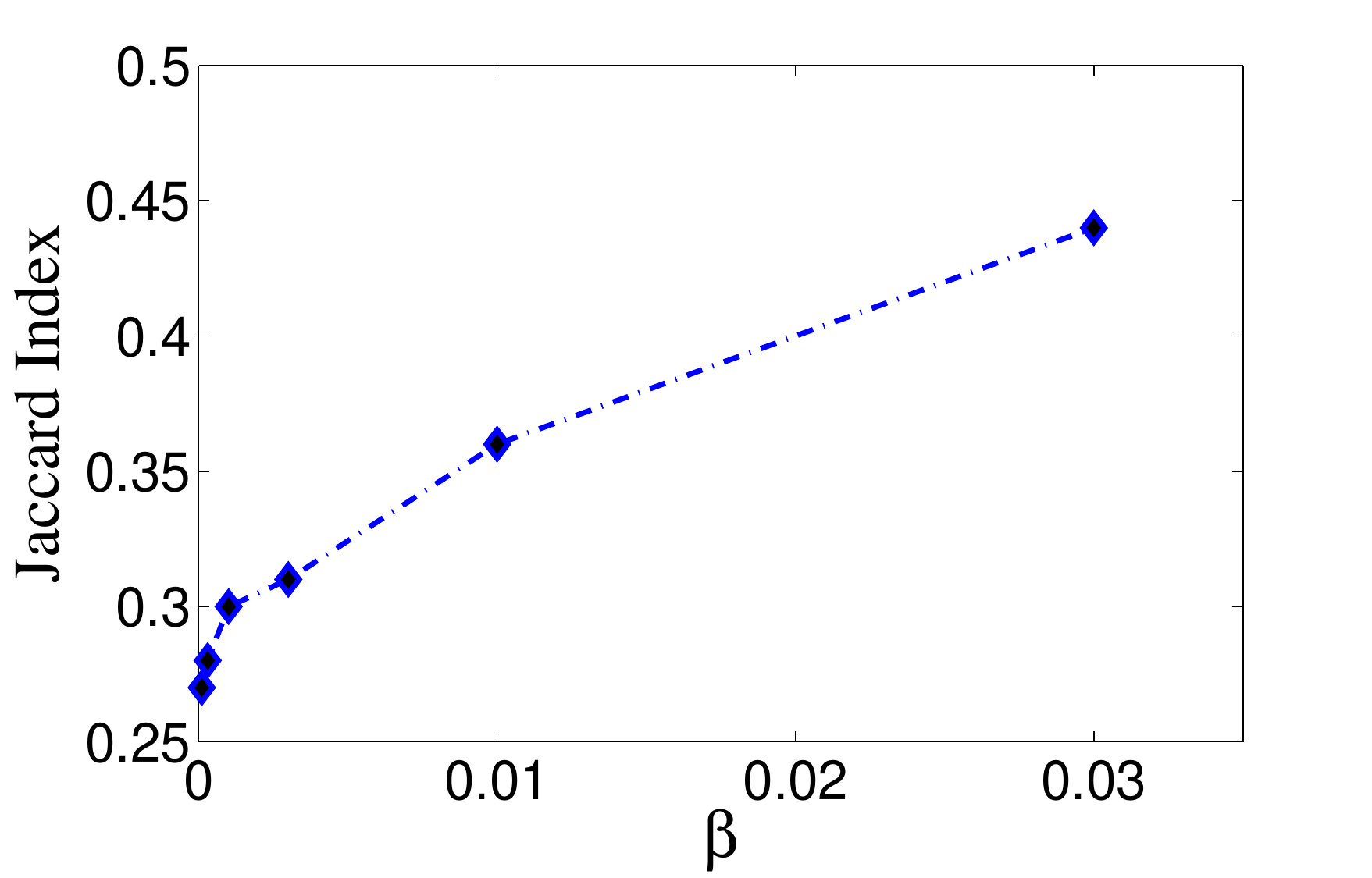}\tabularnewline
(a) & (b) & (c)\tabularnewline
\end{tabular}
\par\end{centering}

\caption{Effect of hyperparameters on (a) discrimination measured in AUC and
(b,c) stability at 90 feature subsets.\label{fig:Effect-of-hyperparameters}}
\end{figure*}

The discrimination of the model with respect to the area under the
ROC curve (AUC) was investigated for various values of the hyperparameters
$\alpha$ and $\beta$ (Fig.~\ref{fig:Effect-of-hyperparameters}a).
The AUC depends more on the lasso hyperparameter $\alpha$, which
controls the number of features being selected. The best AUC for our
model was 0.64 for $\alpha=.004$ and $\beta=.03$. Next, we investigated
the role of the structure hyperparameter $\beta$ on feature stability
against data resampling. For a fixed value of lasso regularization
$\alpha=.004$, increase in $\beta$ resulted in increased stability
of the top 90 features, confirmed by both Consistency index and Jaccard
index (see Figs.~\ref{fig:Effect-of-hyperparameters}b and c). Hence
$\alpha$ affects model AUC, while $\beta$ affects feature stability.

\begin{figure}
\noindent \begin{centering}
\includegraphics[width=0.8\columnwidth]{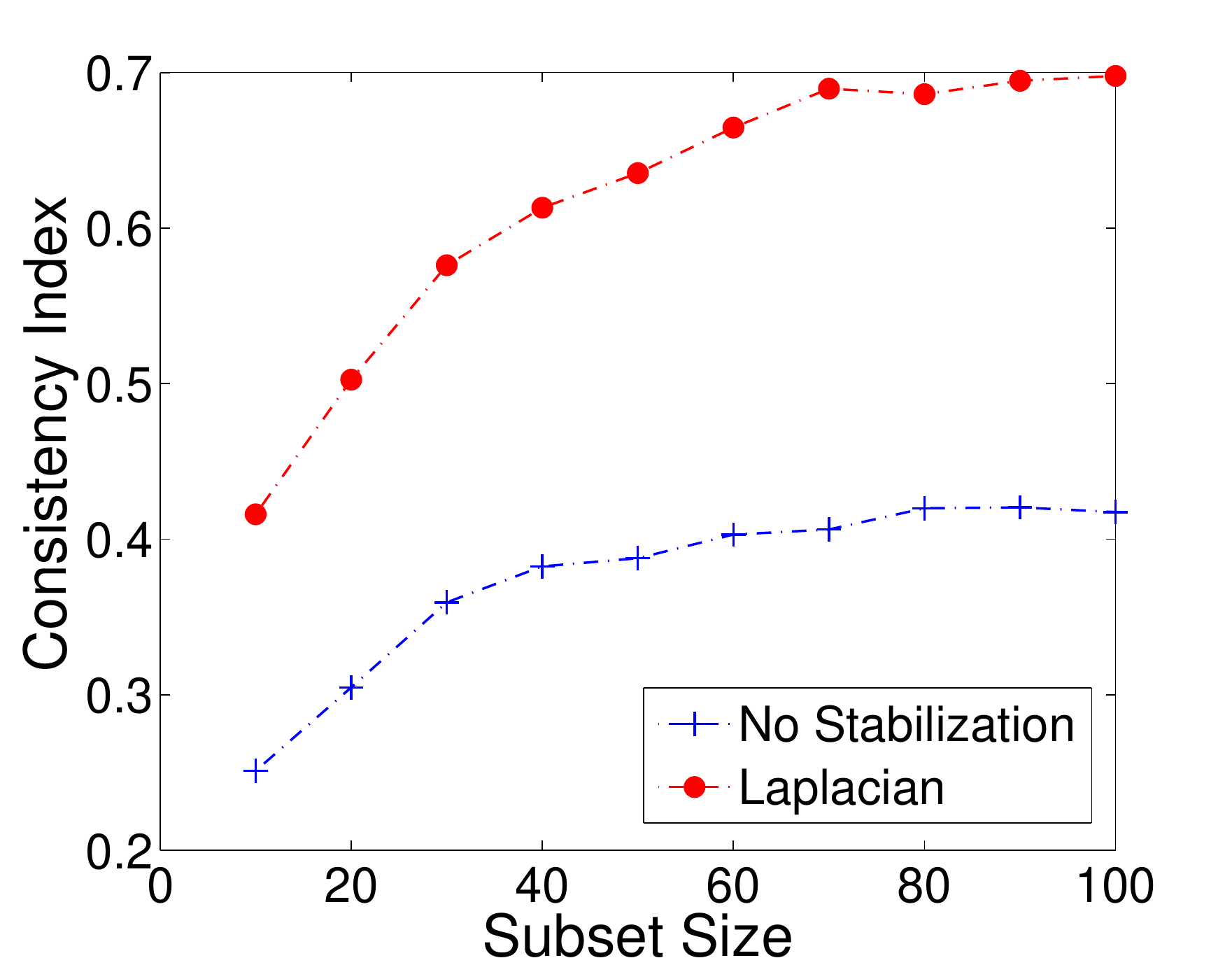}
\par\end{centering}

\caption{Consistency index.\label{fig:Consistency-index.}}
\end{figure}

\begin{figure}
\noindent \begin{centering}
\includegraphics[width=0.8\columnwidth]{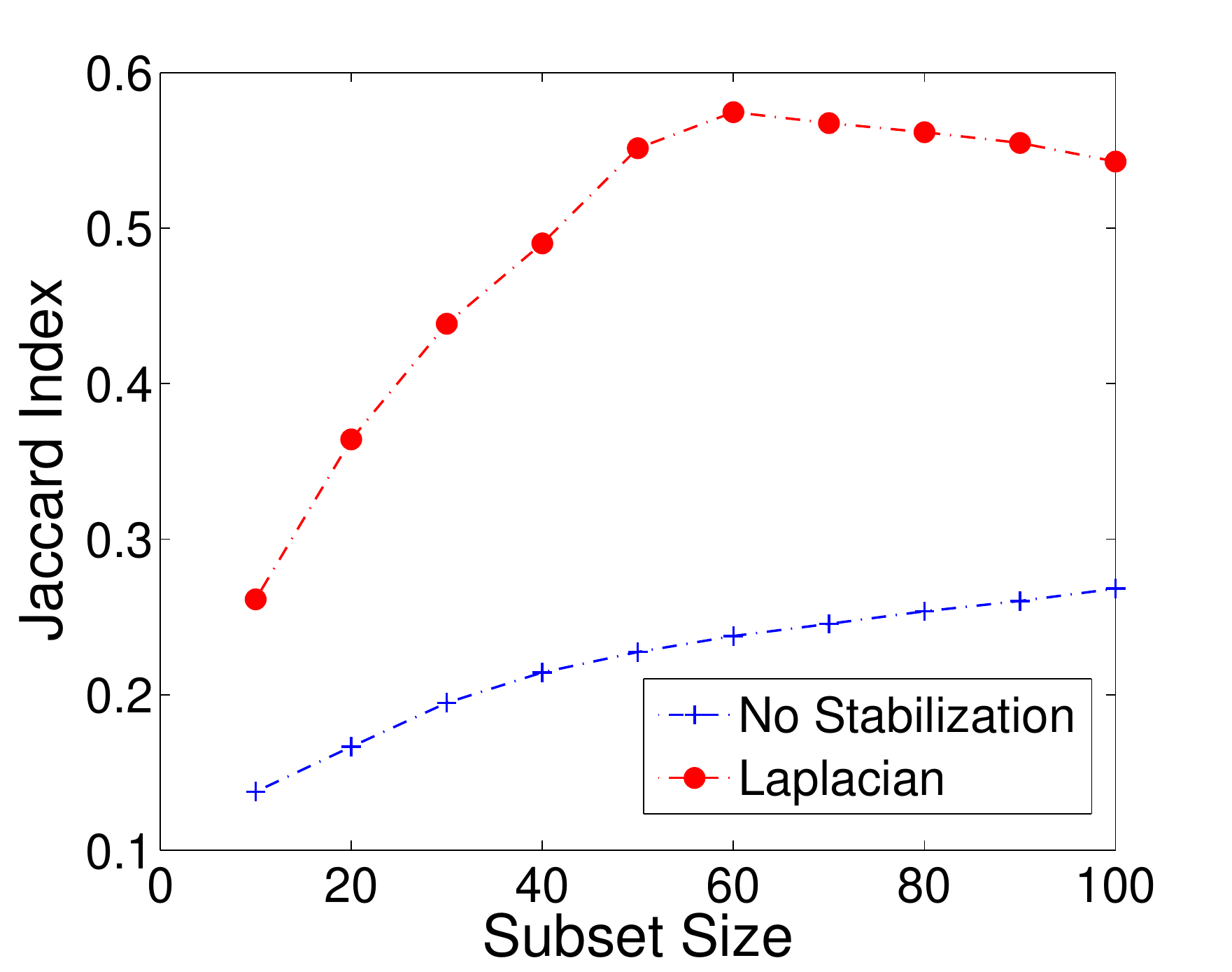}
\par\end{centering}

\caption{Jaccard index.\label{fig:Jaccard-index.}}
\end{figure}

Finally, the stability of feature subsets selected by the unregularized
model and the regularized model with $\alpha=.004$ and $\beta=.03$
were compared for each bootstrap. Our proposed model regularized by
clinical structures was found to be more robust to training data variations
for all subset sizes when measured using both indices (Figs.~\ref{fig:Consistency-index.}
and \ref{fig:Jaccard-index.}).

\begin{table}
\begin{centering}
\caption{The top predictors for heart failure readmission identified by our
proposed model. \label{tab:The-top-predictors}}

\par\end{centering}

\centering{}%
\begin{tabular}{>{\raggedright}p{0.65\columnwidth}r}
\toprule 
\textbf{Top Predictors } & \textbf{Importance}\tabularnewline
\midrule 
Male & 100.0\tabularnewline
\midrule 
Age \textgreater{} 90 & 86.3\tabularnewline
\midrule 
Rare diagnosis in past 3 months & 63.8\tabularnewline
\midrule 
Past respiratory infection  & 55.0\tabularnewline
Disease history in past 6-12 months & 44.9\tabularnewline
Pain in throat and chest  & 41.0\tabularnewline
Chronic kidney disease & 31.7\tabularnewline
Abnormalities of heart beat & 29.3\tabularnewline
Disorders of kidney and ureter & 25.2\tabularnewline
\midrule 
Emerg. admits in past 1-2 years & 48.1\tabularnewline
Admissions in past 2-4 years & 45.7\tabularnewline
Emerg. admits in past 3 months & 39.0\tabularnewline
Emerg-to-ward in 0-3 months & 35.3\tabularnewline
\midrule 
Valvular disease past 3 months & 28.8\tabularnewline
\bottomrule
\end{tabular}
\end{table}

\noindent \begin{center}

\par\end{center}

\noindent \begin{center}

\par\end{center}

\section{Discussion}

Feature stability facilitates reproducibility between model updates
and generalization across medical studies. In this paper, we utilize
the temporal and code structures inherent in EMR data to stabilize
a sparse Cox model for heart failure readmissions. Though heart failure
patients diagnosed with other comorbidities are more prone to rehospitalization
\cite{perkins2013readmission}, our model focuses on patients diagnosed
solely with heart failure. When compared with similar studies, the
model AUC is competitive and the top predictors including male gender,
age, history of prior hospitalizations, presence of kidney disorders
were found to be common \cite{RossJSMulveyGK2008}. Encoding clinical
structures into feature graphs promotes group level selection and
rare-but-important features. This resulted in our model selecting
past rare diagnosis as an important predictor. On two stability measures,
the proposed method has demonstrated to largely improved stability.
 Also, our proposed model is derived entirely from commonly available
data in medical databases. All these factors suggest that our model
could be easily integrated into the clinical pathway to serve as a
fast and inexpensive screening tool in selecting features and patients
for further investigation. Future work includes applying the same
technique for a variety of cohorts and investigating other latent
structures in EMR to enhance feature stability.

\bibliographystyle{latex12}
\bibliography{latex12}

\end{document}